\documentclass[letterpaper, 10 pt, conference, twoside]{ieeeconf} 

\IEEEoverridecommandlockouts                              

\usepackage{amsmath}
\usepackage{amssymb}
\usepackage{mathtools}
\usepackage{graphicx}
\usepackage{float}
\usepackage{array}
\usepackage{tikz}
\usepackage{listings}
\usepackage{hyperref}
\usepackage{graphicx}
\usepackage{cite}
\usepackage{dsfont}
\usepackage{mathtools,etoolbox}
\usepackage[ruled, linesnumbered]{algorithm2e}
\usepackage[none]{hyphenat}
\usepackage[utf8]{inputenc}
\usepackage[english]{babel}

\usepackage{amsthm}
\usepackage{xfrac}
\usepackage{nicefrac}
\usepackage{booktabs}
\usepackage[switch, pagewise]{lineno}
\usepackage{nicefrac}
\usepackage{flushend}

\hypersetup{colorlinks=true,urlcolor=blue,citecolor=red}

\newcommand{\norm}[1]{\left\lVert#1\right\rVert} 

\title{\LARGE \bf MorphEyes: Variable Baseline Stereo For Quadrotor Navigation}
\author{Nitin J. Sanket, Chahat Deep Singh, Varun Asthana, Cornelia Ferm{\"u}ller, Yiannis Aloimonos 
\thanks{This work was supported in parts by the Brin Family Foundation, the Office of Naval Research, the Northrop Grumman Corporation, the National Science Foundation and under Grants N00014-17-1-2622, SMA 1540917 and CNS 1544797 respectively. Varun Asthana was funded by the Maryland Robotics Center's Pathway to PhD program. \textit{(Corresponding author: Nitin J. Sanket.)}}
\thanks{All authors are associated with Perception and Robotics Group, University of Maryland, College Park. Emails: \{{\tt\footnotesize nitin, chahat, vasthana, fer, yiannis}\} {\tt \footnotesize @umiacs.umd.edu}}
\thanks{Digital Object Identifier (DOI): see top of this page.}
}
\begin{document}
\makeatletter
\maketitle
\thispagestyle{plain}
\pagestyle{plain}

\begin{abstract}
Morphable design and depth-based visual control are two upcoming trends leading to advancements in the field of quadrotor autonomy. Stereo-cameras have struck the perfect balance of weight and accuracy of depth estimation but suffer from the problem of depth range being limited and dictated by the baseline chosen at design time. In this paper, we present a framework for quadrotor navigation based on a stereo camera system whose baseline can be adapted on-the-fly. We present a method to calibrate the system at a small number of discrete baselines and interpolate the parameters for the entire baseline range. We present an extensive theoretical analysis of calibration and synchronization errors. We showcase three different applications of such a system for quadrotor navigation: (a) flying through a forest, (b) flying through an unknown shaped/location static/dynamic gap, and (c) accurate  3D pose detection of an independently moving object. We show that our variable baseline system is more accurate and robust in all three scenarios. To our knowledge, this is the first work that applies the concept of morphable design to achieve a variable baseline stereo vision system on a quadrotor.
\end{abstract}


\section*{Supplementary Material}
The supplementary hardware tutorial and
video are available at \url{prg.cs.umd.edu/MorphEyes.html}. 

\begin{figure}[t!]
    \centering
    \includegraphics[width=\columnwidth]{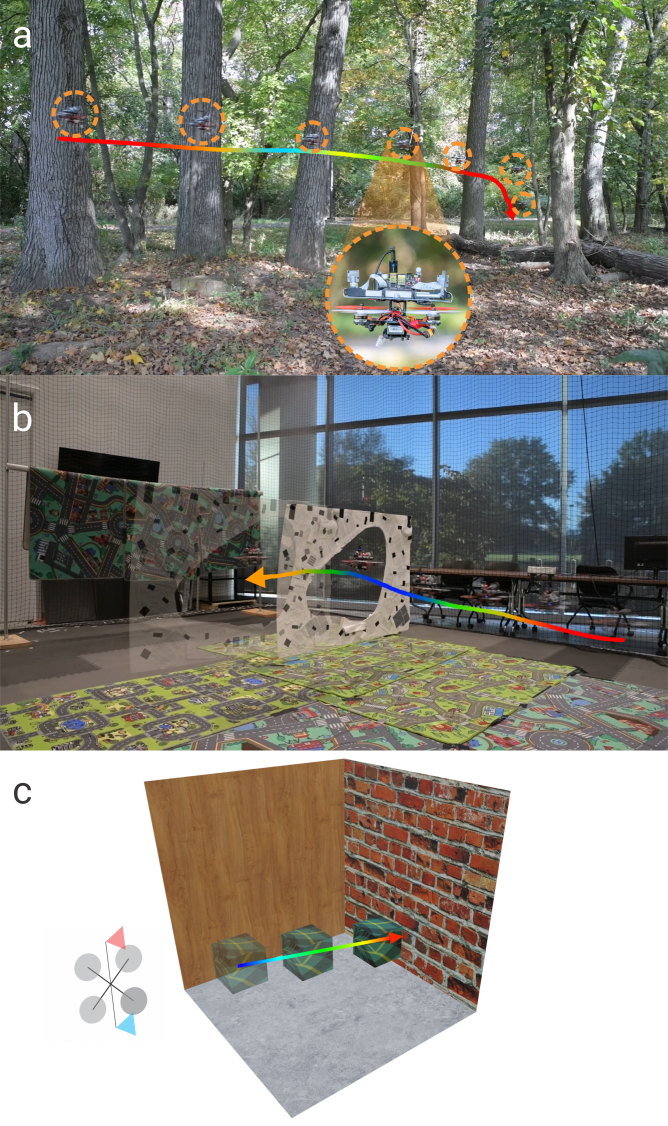}
    \caption{Three applications of a variable baseline stereo system were explored in this work. (a) Flying through a forest, (b) Flying through an unknown shape and location dynamic gap, (c) Detecting an Independently Moving Object. In all the cases, the baseline of the stereo system is changing and is colored coded as jet (blue to red indicates 100 mm to 300 mm baseline). The opacity of the quadrotor/object shows positive progression of time. \textit{All the images in this paper are best viewed in color on a computer.}}
    \label{fig:Banner}
\end{figure}

\section{Introduction}

The rapid rise of autonomous quadrotors in the recent decade has been due to the extensive research in the area of Simultaneous Localization and Mapping (SLAM)\cite{ORBSLAM, RGBDSLAM} and Visual-Inertial Odometry (VIO)\cite{ROVIO, VINS, PRGFlow, SalientDSO} which provide metric depth which can be directly used to perform path planning and thereby achieve navigation. Stereo camera based solutions strike the perfect balance in accuracy, computation power and weight to obtain a depth map when compared to LIDAR or RGB-D systems. However, they suffer from a major drawback that the error grows quadratically with depth, implying that depth in the near-range far exceeds the far-range. Depending on the scenario one might require high accuracy depth in the far range for which the near range resolution has to be increased which results in exorbitant computational cost. To this end, the experts of the field suggested that one could design a stereo system where the baseline could be varied\cite{vbstereo}.

Another trend which is disrupting the quadrotor industry is morphable design which is commonly observed in birds\cite{mintchev2016adaptive}. Such a design philosophy has been applied by works in \cite{mintchev2016adaptive, zhao2017whole, zhao2018design, desbiez2017x, desbiez2017x, zhao2017deformable, falanga2018foldable} to design a quadrotor with morphable frames to traverse narrow gaps or grasp objects. The aforementioned works have systematically modelled the dynamics and developed stable controllers during morph. 

These two upcoming trends, namely, SLAM/Depth estimation and morphable designs have been mutually exclusive and can benefit greatly by being combined. We fill this void by designing a variable baseline stereo system extending the work in \cite{vbstereo} so that we can adapt the baseline for both perceptive and mechanical purposes, which is the first work of such a kind to our knowledge. 

An added benefit of such a design is the usage of the concept of \textit{Active Vision}: one can perform maneuvers and control the image acquisition process, thus introducing new constraints that were not there before \cite{ActiveVision, SukhtameActive, BajcsyActive, GapFlyt}. With our system one doesn't have to move the entire quadrotor which is often expensive in-terms of battery and/or time and can simply move one or more of the cameras to obtain such constraints. 

\subsection{Contributions}
We develop a variable baseline stereo system which estimates disparity and a simple method to obtain extrinsic calibration between the cameras using forward kinematics of a linear actuator. We present an extensive theoretical analysis of our system and three applications for such a system, namely: (a) Navigating through a forest where the baseline changes with scene's minimum depth to keep depth error low at near-obstacles (Fig. \ref{fig:Banner}{\color{red}a}), (b) Flying through an arbitrary shaped small static/dynamic gap by adjusting baseline for better depth estimation and reducing quadrotor footprint (Fig. \ref{fig:Banner}{\color{red}b}) and (c) Improving the accuracy of estimating 3D trajectory of an Independently Moving Object (IMO) by adjusting baseline (Fig. \ref{fig:Banner}{\color{red}c}). 

\subsection{Organization of the paper}
We present an extensive theoretical analysis of errors in a variable baseline stereo system in Sec. \ref{sec:TheoAna}. We talk about the software and hardware design of our system in Sec. \ref{sec:HWSetup}. Three different applications of the proposed system are discussed in Sec. \ref{sec:Expts} and we conclude the paper in Sec. \ref{sec:Conc}.

\section{Theoretical Analysis}
\label{sec:TheoAna}
Let $\epsilon_z$, $\epsilon_d$, $b$, $f$ and $z$ denote the depth error, disparity error, baseline, focal length and depth respectively. They are related by\cite{vbstereo} 
\[
\epsilon_z = \frac{z^2}{bf}\epsilon_d
\]
In the above expression, we assume that $b$ is obtained after both camera feeds are distortion corrected, stereo-rectified and perfectly time synchronized. These corrections rely on both intrinsic and extrinsic camera calibration which can affect the disparity error $\epsilon_d$ and in-turn affect the depth error $\epsilon_z$. We discuss this next.

Let $K$ denote the intrinsic camera calibration matrix \cite{Bouguet}:

\[ K = \begin{bmatrix} 
f_x & \alpha f_x & c_x\\
0 & f_y & c_y\\
0 & 0 & 1\\
\end{bmatrix}
\]

Here, $f_x$, $f_y$, $c =  \begin{bmatrix}c_x & c_y \end{bmatrix}^T$ and $\alpha$ denote the focal length in $x$, $y$, principle point and skew respectively. We'll use a subscript $L$ or $R$ to denote left and right cameras (assuming horizontal stereo system) respectively whenever required. We also denote the extrinsics between the left and right cameras (left camera being origin) as $\{R, T\}$. 

The first step in a stereo pipeline is to undistort/distrotion rectify both feeds. Commonly, a radial-tangential model is used for distortion correction \cite{heikkila1997four, tang2017precision, zhang1999flexible} which is given as follows

\begin{align}
\mathbf{x}_r &= \left( 1 + k_1r^2 + k_2r^4 + k_5r^6\right)\mathbf{x} + \begin{bmatrix} 
2k_3xy + k_4\left(r^2 + 2x^2 \right) \\
k_3\left(r^2 + 2y^2\right) + 2k_4xy\\
\end{bmatrix}\\
\mathbf{x}_t &= \begin{bmatrix} 
f_x\left(x_r + \alpha y_r\right) + c_x\\
f_yy_r + c_y\\
\end{bmatrix}; \quad r = \sqrt{x^2 + y^2}=\norm{\mathbf{x}}_2
\end{align}

Where $\mathbf{x}_t = \begin{bmatrix} x_t & y_t\end{bmatrix}^T$, $\mathbf{x} = \begin{bmatrix} x & y\end{bmatrix}^T$, $k_1, \cdots, k_5$ denote the corrected co-ordinates, original co-ordinates and distortion parameters respectively. 

Once this distortion correction has been performed, next step is stereo-rectification, i.e., transformation to make epipolar lines parallel in a stereo camera. This is done by relative rotation removal as follows \cite{Bouguet}
\begin{align}
\mathbf{\tilde{x}}_L &= K_LR_\text{rect}K_L^{-1}\mathbf{x}_{t,L}\\
\mathbf{\tilde{x}}_R &= K_RR_\text{rect}R^TK_R^{-1}\mathbf{x}_{t,R}
\end{align}

where, $K_LR_\text{rect}K_L^{-1}$ and $K_RR_\text{rect}R^TK_R^{-1}$ are generally denoted as $H_L$ and $H_R$ respectively denoting the pure rotation homographies for stereo-rectification. $R_\text{rect}$ is given as follows
\begin{align}
R_\text{rect} &= \begin{bmatrix} r_x^T & r_y^T & r_z^T\end{bmatrix}^T; \,\, \tilde{r}_z = \begin{bmatrix} 0 & 0 & 1 \end{bmatrix}^T \\
r_x = \frac{T}{\norm{T}}; \,\, r_z &= \frac{\tilde{r}_z - \left(\tilde{r}_z\cdot r_x\right)r_x}{\norm{\tilde{r}_z - \left(\tilde{r}_z\cdot r_x\right)r_x}_2}; \,\, r_y = r_z \times r_x
\end{align}

In practice, each of the estimated quantities $\{R, T\}, k_{1,L/R}, \cdots, k_{5,L/R}, \alpha_{L/R}, f_{x/y,L/R}$ have errors. Let us denote percentage additive error in a quantity $\hat{a}$ as $a_e$, such that the estimate $\tilde{a} = \hat{a}(1 + (a_e/100))$ (for rotation matrix $R$, if $R_e$ is an error rotation matrix, the estimate is defined as $\tilde{R}=\hat{R}R_e$). Also, due to imperfect undistortion or stereo-rectification a pixel which should ideally be at $\mathbf{\hat{x}}$ falls at $\mathbf{\tilde{x}}$ giving rise to error in both $x$ and $y$ directions which we denote as $e_x = \norm{\hat{x} - \tilde{x}}_1$ and $e_y = \norm{\hat{y} - \tilde{y}}_1$ respectively.

\begin{figure}[t!]
    \centering
    \includegraphics[width=\columnwidth]{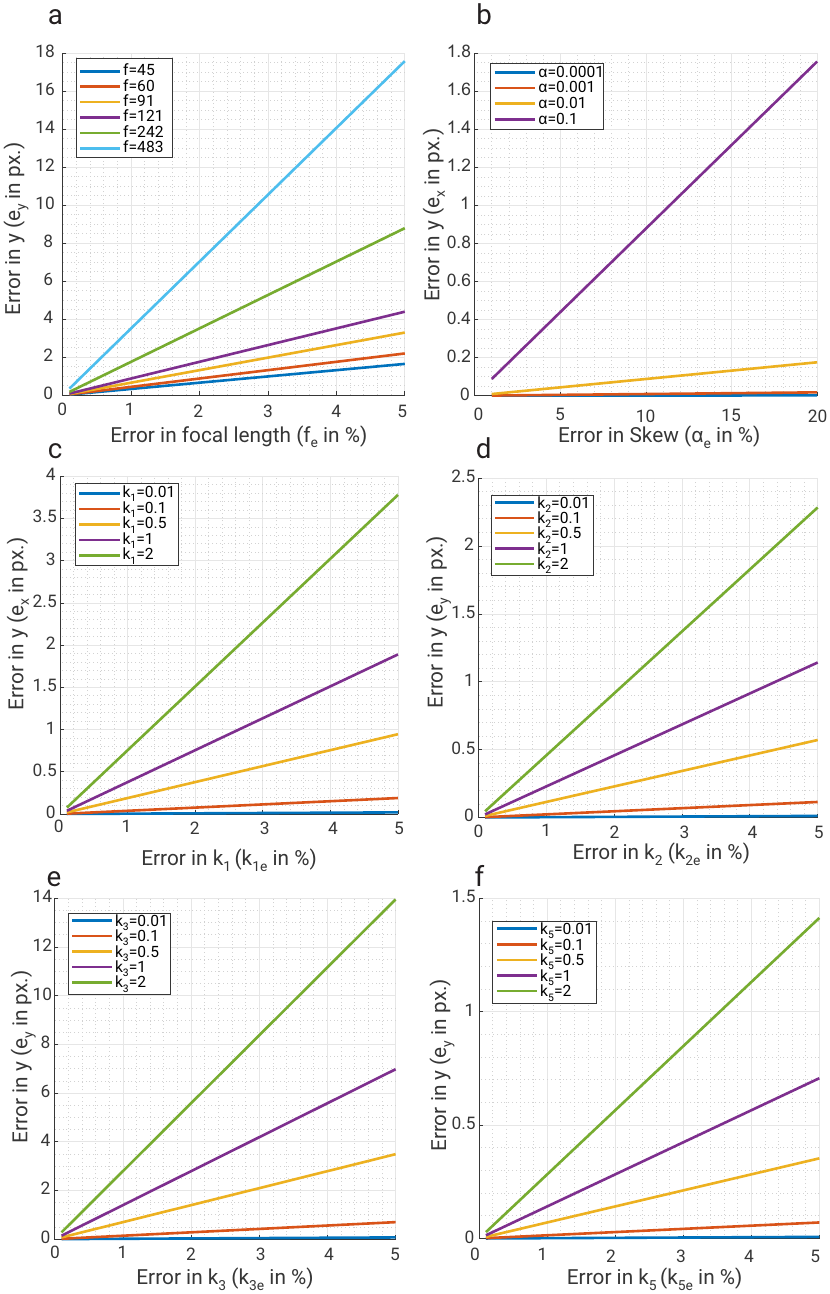}
    \caption{Error in pixel location due to error in various estimated intrinsic parameters. (a) $e_y$ vs. $f_e$, (b) $e_x$ vs. $\alpha_e$, (c) $e_x$ vs. $k_{1e}$, (d) $e_y$ vs. $k_{2e}$, (e) $e_y$ vs. $k_{3e}$, (3) $e_y$ vs. $k_{5e}$. \textit{Notice that the $X$ and $Y$ scales for each of the plots is different though trend may seem similar.}}
    \label{fig:IntrinsicErrors}
\end{figure}
\begin{figure}[t!]
    \centering
    \includegraphics[width=\columnwidth]{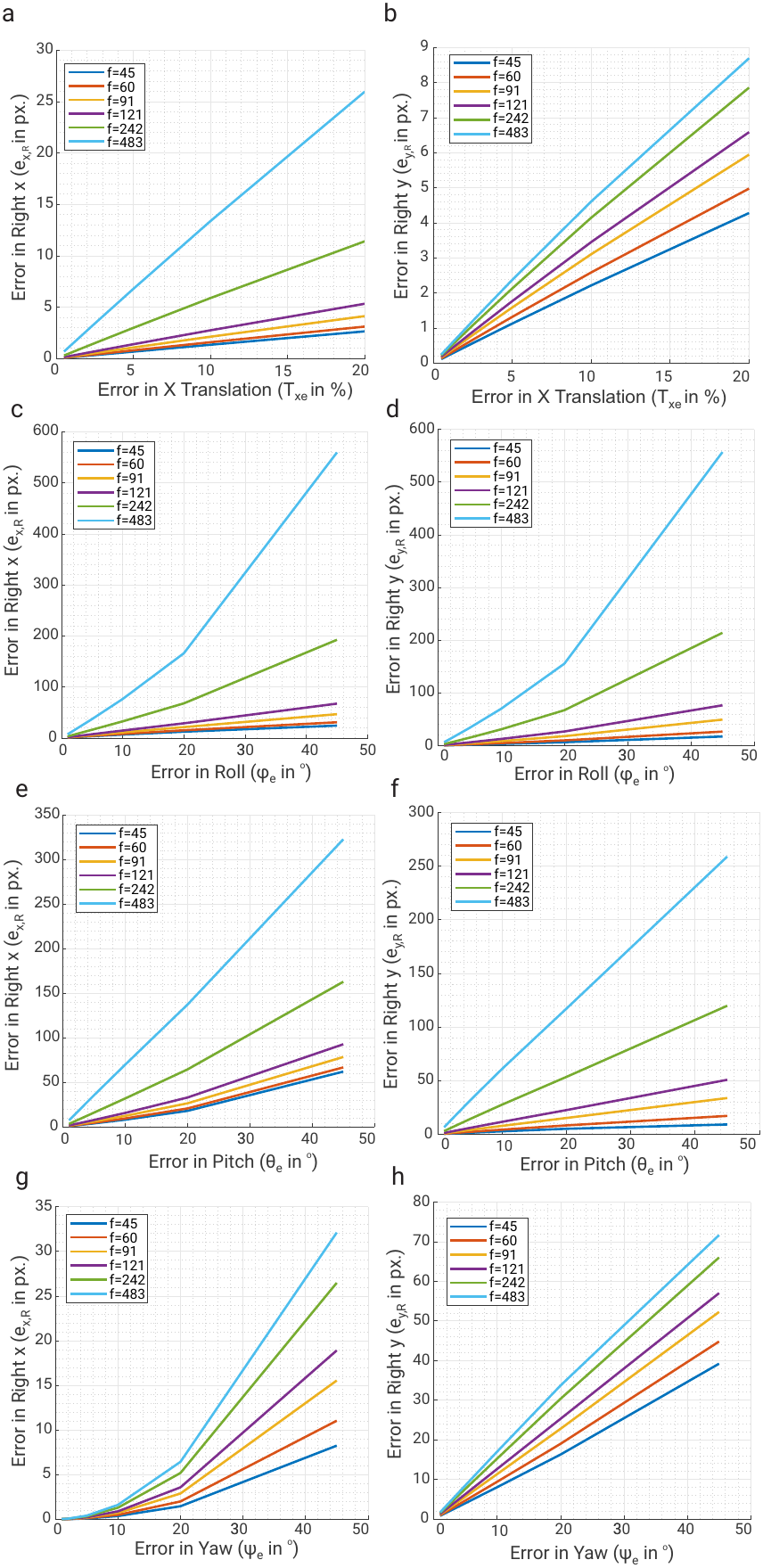}
    \caption{Error in pixel location in right camera due to error in various estimated extrinsic parameters. (a) $e_{x,R}$ vs. $T_{xe}$, (b) $e_{y,R}$ vs. $T_{xe}$,
    (c) $e_{x,R}$ vs. $\phi_e$, (d) $e_{y,R}$ vs. $\phi_e$, (e) $e_{x,R}$ vs. $\theta_{e}$, (f) $e_{y,R}$ vs. $\theta_{e}$, (g) $e_{x,R}$ vs. $\psi_e$, (h) $e_{y,R}$ vs. $\psi_e$. \textit{Notice that the $X$ and $Y$ scales for each of the plots is different though trend may seem similar.}}
    \label{fig:ExtrinsicErrors}
\end{figure}

The interaction of each estimated quantity is complex to analyse by merely looking at the equations. Hence, to provide intuition we visualize the variation of each parameter and how it affects the error. Figs. \ref{fig:IntrinsicErrors} and \ref{fig:ExtrinsicErrors} show the variation of $e_x$ and $e_y$ with different intrinsic and extrinsic quantities respectively. \textit{Notice that the $X$ and $Y$ scales for each of the plots is different though trend may seem similar.}

For each plot we assume the error is only due to one parameter. For the aforementioned analysis an image size of 128$\times$128 px. is used with a sensor size of 4.24 mm (1/3" square sensor commonly used for robotics) and focal length of [1.5, 2, 3, 4, 8, 16] mm. 

In practice, the errors of all the factors are combined. The relationship between $x$ and $y$ errors for variation of each parameter is summarized in Table \ref{tab:exeyIntrinsicAndExtrinsic}. 

\begin{figure}[t!]
    \centering
    \includegraphics[width=0.5\columnwidth]{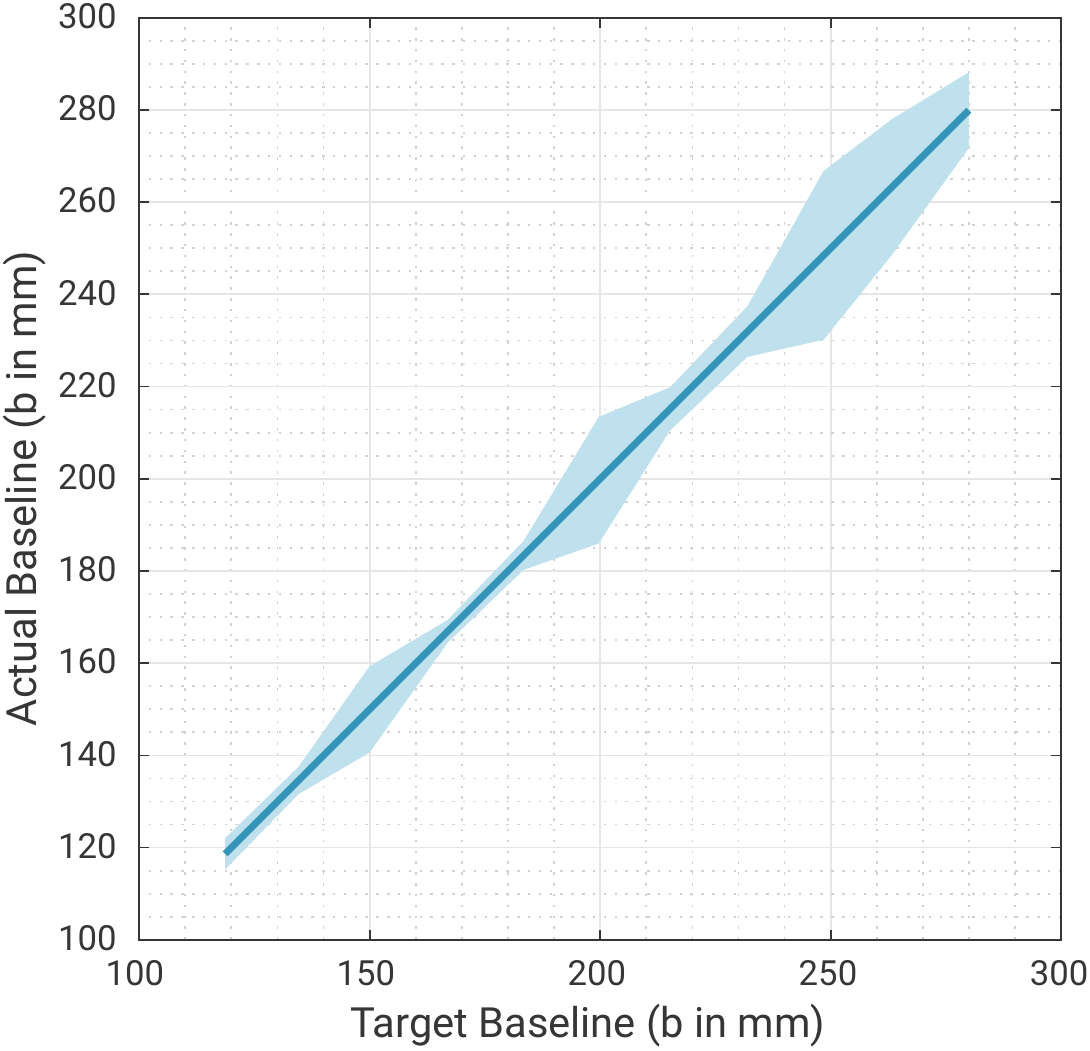}
    \caption{Target vs. achieved baseline. \textit{The highlight shows the 10$\sigma$ value.}}
    \label{fig:BaselineCalib}
\end{figure}

A plot of target versus achieved baseline is shown in Fig. \ref{fig:BaselineCalib} where the mean of 10 samples for each position is plotted as a dark line with 10$\sigma$ value shown as highlight. Notice that we achieve almost a perfect straight line of slope 1. The maximum calibration error of 1.83 mm was observed which is about 0.7\% of the baseline.  

\begin{figure}[t!]
    \centering
    \includegraphics[width=0.5\columnwidth]{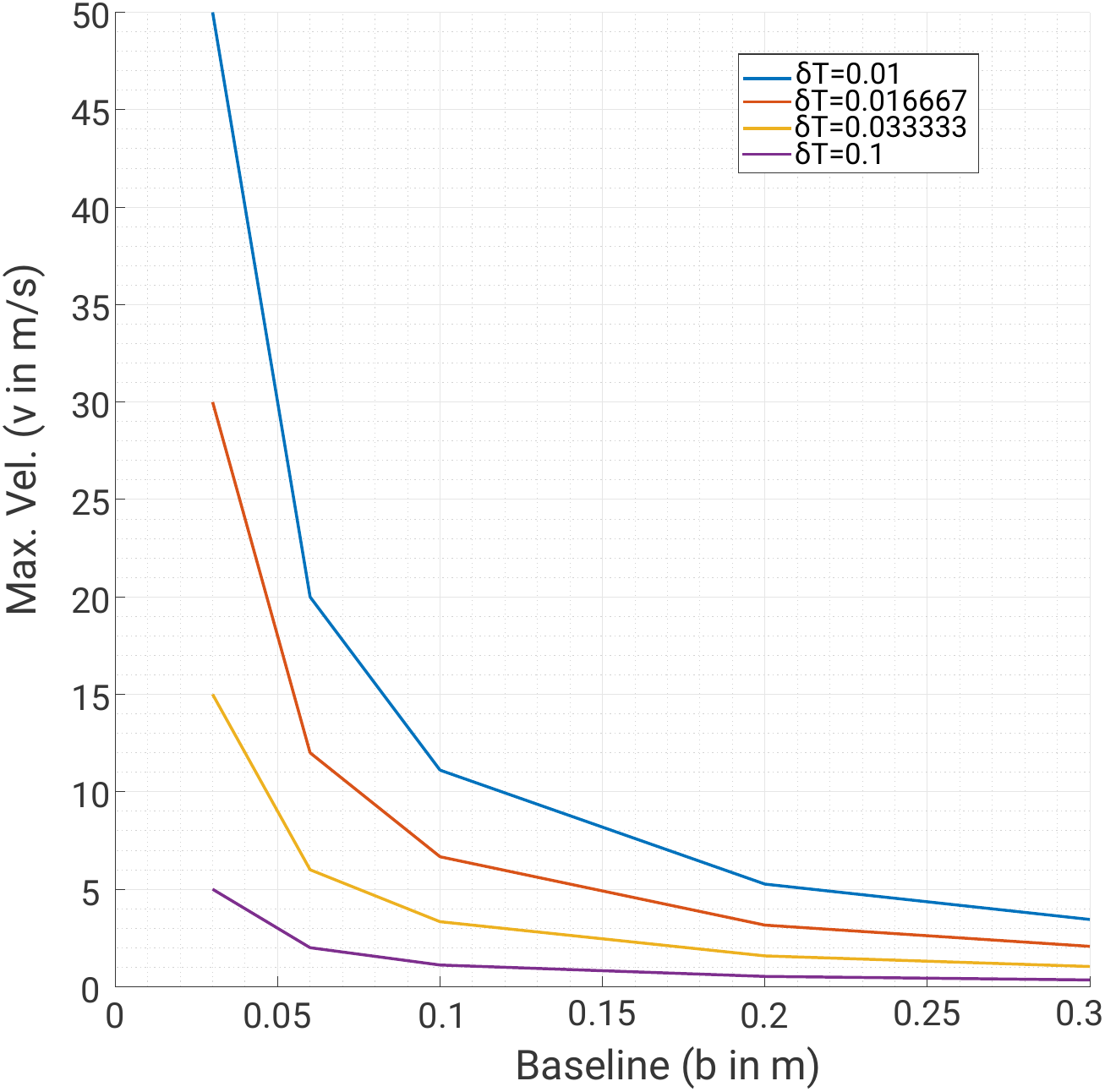}
    \caption{Max. velocity to have a disparity error lower than $k$ px. vs. baseline for different time synchronization errors ($\delta t$).}
    \label{fig:SyncErrors}
\end{figure}

\begin{figure}[t!]
    \centering
    \includegraphics[width=0.8\columnwidth]{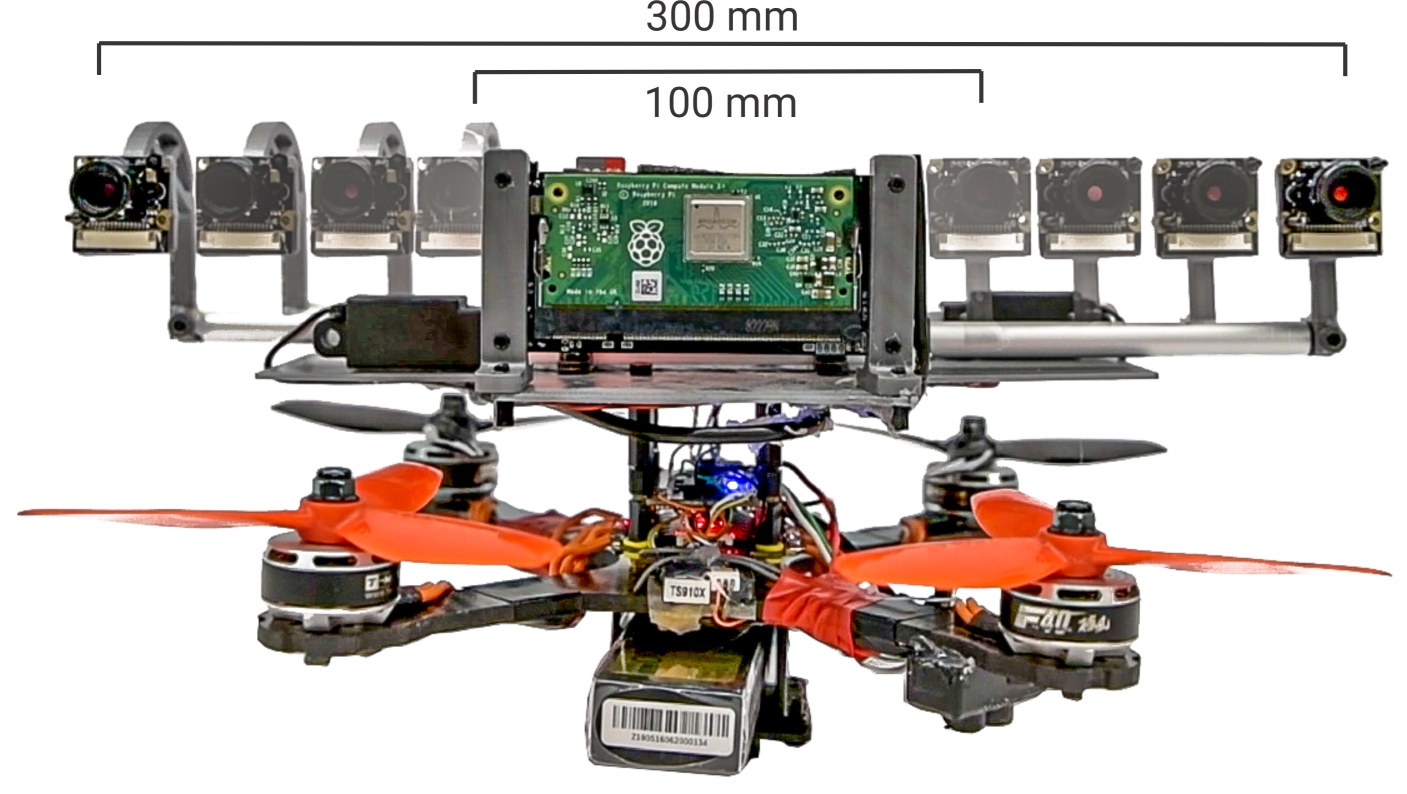}
    \caption{Variation of baseline from 100 mm to 300 mm. Notice that the stereo system is bigger than the quadrotor at the largest baseline.}
    \label{fig:VariableBaseLineQuad}
\end{figure}

\begin{figure}[t!]
    \centering
    \includegraphics[width=\columnwidth]{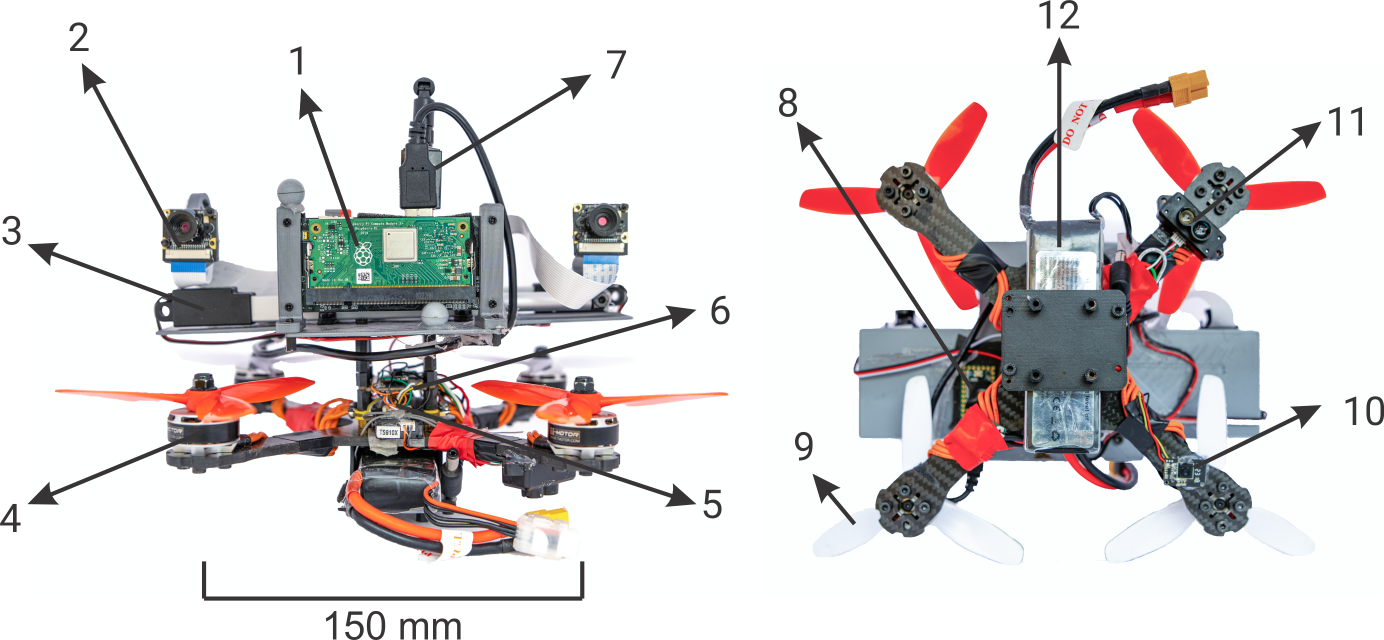}
    \caption{Quadrotor platform used for experiments. (1) RaspberryPi 3B+ compute module, (2) Stereo camera, (3) Actuonix linear servo, (4) T-Motor F40 III Motors, (5) T-Motor F55A 4-in-1 ESC, (6) Holybro Kakute F7 flight controller, (7) WiFi module, (8) Teensy 3.2 microcontroller, (9) 5045$\times$3 propeller, (10) Optical Flow module, (11) TFMini lidar, (12) 3S LiPo battery.}
    \label{fig:QuadParts}
\end{figure}

\begin{figure*}[t!]
    \centering
    \includegraphics[width=0.8\textwidth]{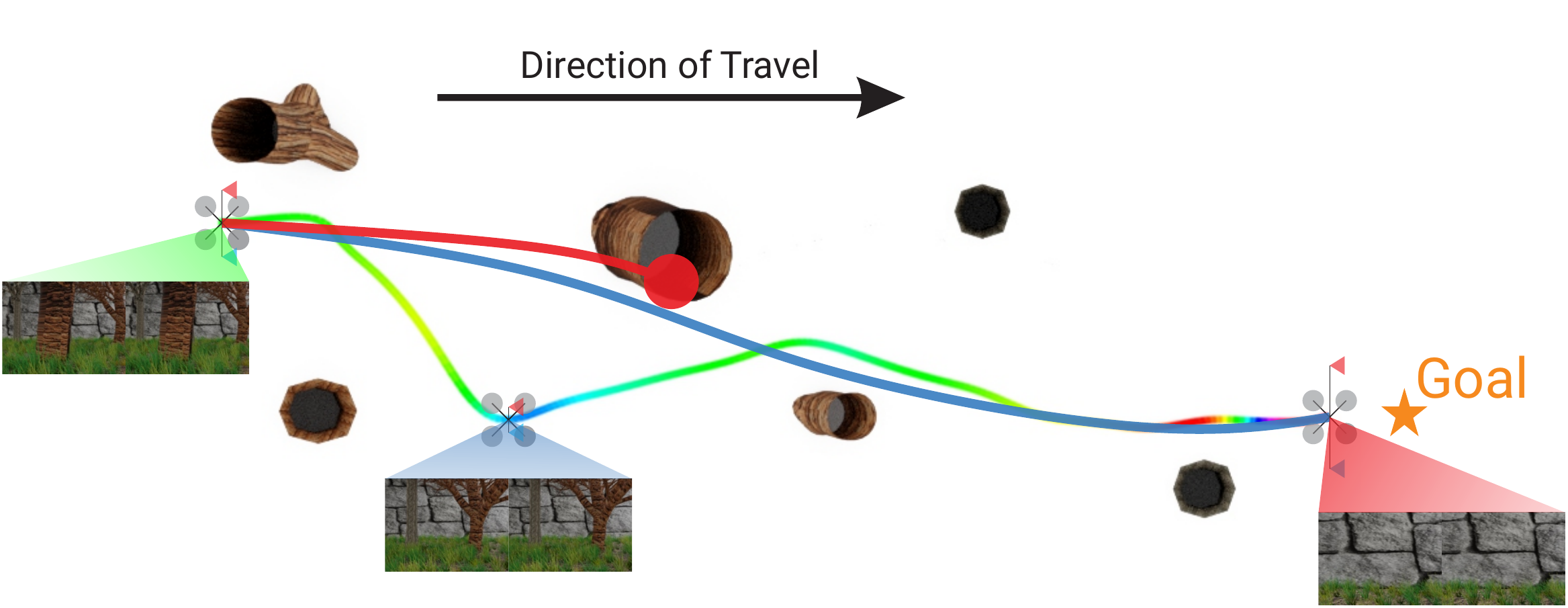}
    \caption{Variable baseline stereo performance in simulated forest flight when compared to small and large baselines. Note that the large baseline system crashes (red curve) and small baseline system (blue curve) can traverse the scene but is about $4\times$ slower than the variable baseline system. The baseline for the variable baseline case is color-coded as jet (blue to red indicates small to large baseline).}
    \label{fig:BlenderTopSceneForrest}
\end{figure*}

\begin{figure*}[t!]
    \centering
    \includegraphics[width=\textwidth]{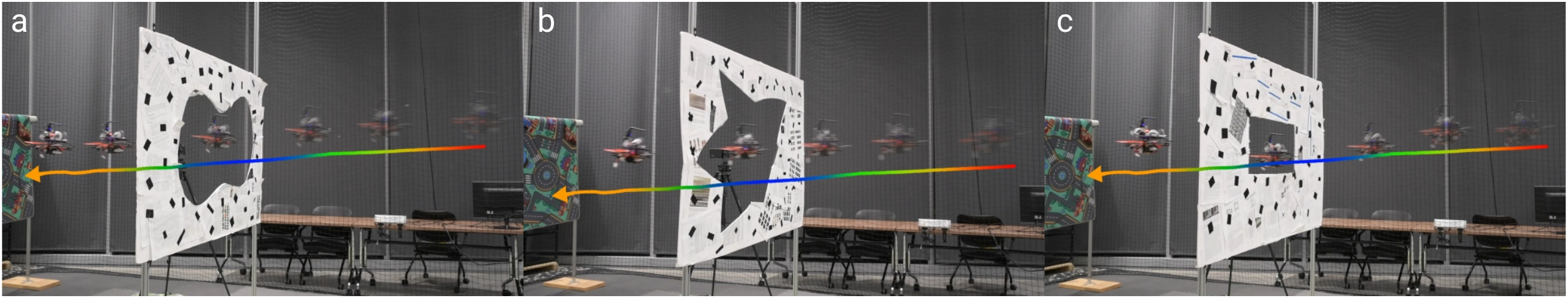}
    \caption{Sequence of images of quadrotor going through different shaped gaps: (a) Infinity, (b) Goku, (c) Rectangle. In all the cases, the baseline of the stereo system is changing and is colored coded as jet (blue to red indicates 100 mm to 300 mm baseline).}
    \label{fig:GapFlytStatic}
\end{figure*}

\begin{figure}[t!]
    \centering
    \includegraphics[width=\columnwidth]{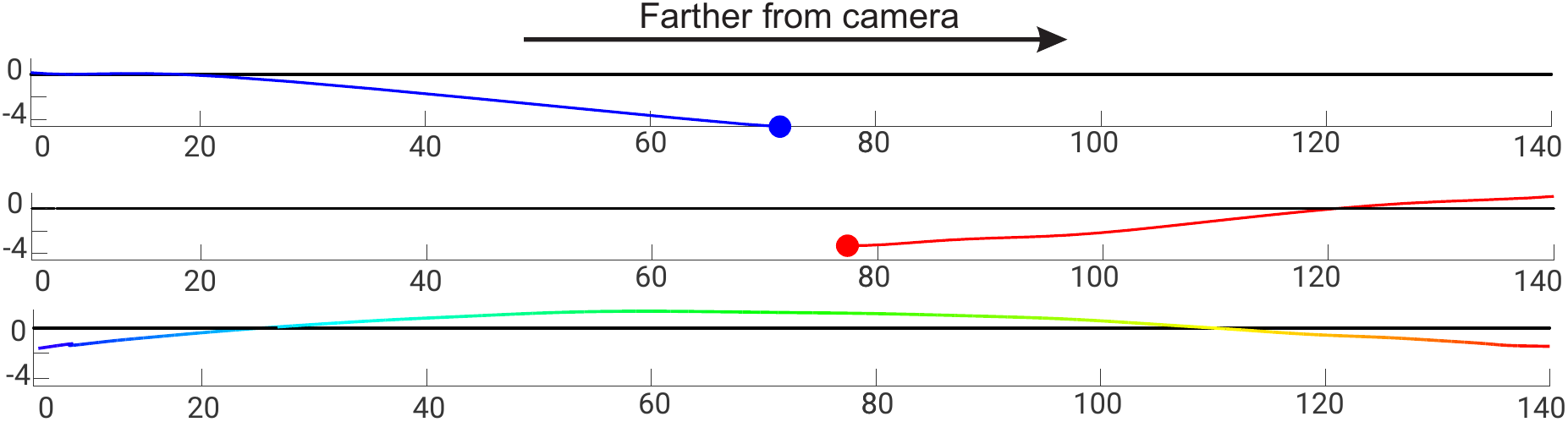}
    \caption{Variable baseline stereo performance in simulated 3D IMO detection when compared to small and large baselines. Note that both the large baseline system (red curve) and small baseline system (blue curve) lose detection of the IMO at different parts of the scene (dots). The baseline for the variable baseline case is color-coded as jet (blue to red indicates small to large baseline). Black curve (horizontal line at zero vertical axis) represents the ground truth trajectory of the object.}
    \label{fig:IMODiffBaselines}
\end{figure}

\begin{table}[t!]
\centering
\caption{Relationship between $e_x$ and $e_y$ for errors in different parameters.}
\resizebox{0.7\columnwidth}{!}{
\label{tab:exeyIntrinsicAndExtrinsic}
\begin{tabular}{lllllll}
\toprule
\multicolumn{7}{c}{Intrinsic parameters}\\
\hline
& $f_e$ & $\alpha_e$ & $k_{1e}$ & $k_{2e}$ & $k_{3e}^*$ & $k_{5e}$ \\
\hline
$e_y/e_x$ & $\approx 1$ & 0 & $\approx 1$ & $\approx 1$ & $\approx 2$& $\approx 1$\\
\hline
\multicolumn{7}{c}{Extrinsic parameters}\\
\hline
& $T_{xe}^\dagger$ & $\phi_e$ & $\theta_{e}$ & $\psi_{e}$ & & \\
\hline
$e_{x,L}/e_{x,R}$ & 1 & 0 & 0 & 0 & & \\
$e_{y,L}/e_{y,R}$ & 1 & 0 & 0 & 0 & & \\
\hline
 \bottomrule
\end{tabular}}
  \\\tiny{$^*$ $k_{4e}$ and $k_{3e}$ are related, i.e., $e_x$ for $k_{4e}$ = $e_y$ for $k_{3e}$ and vice-versa. $^\dagger$ Errors for $T_{ye}$ and $T_{ze}$ are same as $T_{xe}$.}
\end{table}

The error in the axis parallel to the baseline of the stereo-camera (image X-axis for a horizontal stereo system and image Y-axis for a vertical stereo system) due to errors in intrinsics (including lens distortion) and extrinsics (stereo-rectification) is important only when the error magnitude is significantly larger than the search window (in-case of classical block matching). In such a case this error is directly related since the disparity error $\epsilon_d$ as it would lead to false matches (assuming zero matching error or perfect matching). The error in the axis normal to the baseline axis has a more complex pattern and depends on the structure of the scene, i.e., if a pixel is shifted up/down (in-case of horizontal stereo system), the pixel could land at the edge of object boundaries giving a huge disparity error $\epsilon_d$ (since it is estimated by matching along the baseline axis) \cite{fermuller1997visual, cheong1998effects}. 

Off-synchronization between the stereo cameras can also produce false disparity. The criterion for a disparity error (first order Taylor series approximation) of $k$ px. when the camera is moving with a velocity of $v$ ms$^{-1}$ (acceleration of zero) and an synchronization error time of $\delta t$ for a scene at depth $Z$ is given by

\[
v\delta t \ge \frac{kZ^2}{bf - kZ}
\]

Fig. \ref{fig:SyncErrors} shows the maximum achievable velocity to keep disparity error under $k=1$ px. for $Z=1$ m. Note that, in our work we do not vary the resolution of the sensor and only vary the baseline to increase far-range depth accuracy with the loss of near-range depth accuracy. A detailed analysis of this can be found in the fixed-resolution stereo section of \cite{vbstereo}.  


\section{Hardware and Software Design}
\label{sec:HWSetup}
\subsection{Hardware Setup}
Our variable baseline stereo system is built using a StereoPi\footnote{\url{https://stereopi.com/}} and Actuonix L12-R linear servos\footnote{\url{https://www.actuonix.com/L12-R-Linear-Servo-For-Radio-Control-p/l12-r.htm}}. The StereoPi is a stereo camera system based on the RaspberryPi 3B+ coupled to two waveshare cameras with a Omnivision OV5647 sensor mated to a 3.96 mm f/2.6 lens with a diagonal field of view of 56.8$^\circ$. The Actunonix linear servo used has a stroke length of 100 mm and a gear ratio of 50:1 (Fig. \ref{fig:VariableBaseLineQuad} shows the system with different baselines from 100 mm to 300 mm). These components were chosen for their ease of use and cost-effectiveness for the performance obtained. They can easily be replaced with counterparts from other manufacturers.   

\subsection{Software Stack}
Our software stack includes three sub-modules: (a) Camera drivers and disparity estimation, (b) Camera calibration prediction, and (c) Control of camera baseline. Each sub-module is discussed next.

The camera drivers are implemented in C++ which are Robot Operating System (ROS) compatible. Disparity is obtained using a simplified version of \cite{hirschmuller2007stereo}.

The cameras were calibrated using the \textsc{Matlab}'s\footnote{\url{https://www.mathworks.com/products/matlab.html}} camera calibration toolbox using images of the checkerboard on ten equally spaced baselines between 100 and 300 mm. The extrinsics are represented as a dual-quaternion \cite{jia2013dual, daniilidis1999hand}. We interpolate between extrinsics on-the-fly (for intermediate values) using the screw linear interpolation (ScLERP)  \cite{shoemake1985animating} for its speed and accuracy since the motion of the cameras are linear. 

Finally, we control the camera baseline using ROS communicating with a Teensy 3.2 microcontroller which in-turn controls the servos. The camera baseline control methodology depends on the application and is explained in Secs. \ref{subsec:ExptsForest}, \ref{subsec:ExptsGapFlyt} and \ref{subsec:ExptsIMO}.

\section{Experiments: Applications}
\label{sec:Expts}
\subsection{Quadrotor Platform}
The quadrotor used in the experiments is a custom-built platform called PRGCorgi210$\alpha$\footnote{\url{https://github.com/prgumd/PRGFlyt/wiki/PRGCorgi}} (Fig. \ref{fig:QuadParts}). The platform is built on a X-shaped 210 mm sized (motor to motor dimension) racing frame. The motors used are T-Motor F40III KV2400 mated to 5040$\times$3 propellers. The lower level controller and position hold is handled by ArduCopter 4.0.4dev firmware running on the Holybro Kakute F7 flight controller mated to an optical flow sensor and TFMini LIDAR as altimeter source. All the higher level navigational commands are sent by the companion computer (RaspberryPi) using RC-Override to the flight controller running in Loiter mode using MAVROS. The RaspberryPi runs all the vision and planning algorithms on-board at 1 Hz. The quadrotor take-off weight including the battery is 780 g and has a thrust to weight ratio of 2:1. 

\subsection{Simulation Environment}
We rendered stereo camera images with variable baseline using Blender$^\text{\textregistered}$\footnote{\url{https://www.blender.org/}} 3D creation software. We modelled two scenes for different experiments: (a) Stereo Camera Navigating through a Forest (Fig. \ref{fig:BlenderTopSceneForrest}) with a variety of trees and grass, and (b) a room with a cube moving away from the stereo camera (Fig. \ref{fig:Banner}{\color{red}c}). The camera baseline changes based on the depth of the scene which is estimated using \cite{hirschmuller2007stereo} (Secs. \ref{subsec:ExptsForest}, \ref{subsec:ExptsGapFlyt} and \ref{subsec:ExptsIMO}). The images are rendered in 640 $\times$ 480 px. resolution using Cycles render engine.  

As mentioned before, we present three applications of our a variable baseline stereo system for quadrotor navigation which are explained in the next three sub-sections.

\subsection{Forest Navigation}
\label{subsec:ExptsForest}
In this scenario, we are tasked with the problem of navigating through a forest. The aim is to navigate to the goal location without any collisions. We compute per-pixel depth from the disparity obtained using $Z = \frac{bf}{d}$. The control policy is used to change the current heading direction (velocity vector) using a simple Proportional-Integrative-Derivative (PID) controller to reach the goal whilst avoiding collisions. The current desired direction vector $\tilde{\mathbf{v}}_g$ is given by a weighted sum of the goal direction $\mathbf{v}_g$ and free path direction $\mathbf{v}_\text{free})$. The goal direction can be treated as a global path planner and the free path direction vector can be treated as a local path planner. The intuition is that when there is free space the global planner takes a larger weight and vice-versa to avoid collisions whilst still going towards the goal. Note that this strategy is not necessarily optimal and other strategies \cite{zhou2019robust, liu2017search} can be used to achieve optimality which require building of a map which is generally expensive. 

The free path direction $\mathbf{v}_\text{free}$ is obtained as follows: Consider a small neighborhood $\mathcal{N}$ on the image plane centered around where the goal direction vector intersects the image plane (this could be outside the visible region in which case the edge of the frame closest to the goal direction is chosen). $\mathbf{v}_\text{free}$ is chosen as the center of the largest free space in the neighborhood $\mathcal{N}$. 

Now, let $Z_\text{close}$ denote the closest depth value in this neighborhood. 
The neighborhood $\mathcal{N}$ is chosen to minimize the control effort and to avoid large changes in velocity (accelerations) and can be thought of as the projection of the quadrotor with safety margin onto the depth image. 
The final control equations are given by

\begin{align}
\tilde{\mathbf{v}}_g &= (1-w)\mathbf{v}_g + w\mathbf{v}_\text{free}; \,\, w\in[0,1]\\
\mathbf{e}(t) &= \tilde{\mathbf{v}}_g(t) - \tilde{\mathbf{v}}_\text{curr}(t)\\
\mathbf{u}(t) &= K_p\mathbf{e}(t) + K_i\int_0^\tau\mathbf{e}(\tau)d\tau +  K_d\frac{d\mathbf{e}(t)}{dt}
\\
w &= \frac{1}{1+e^\frac{-1}{Z_\text{close}}}; \,\, Z_\text{close} = \min{Z(x,y)\,\,\forall\,\,\{x, y\} \subset \mathcal{N}}
\end{align}

where $\tilde{\mathbf{v}}_\text{curr}(t)$ is the estimated current heading direction. Notice how the current/local goal vector gets dominated by $\mathbf{v}_\text{free})$ when we get close to a collision ($w\rightarrow 1$) and by $\mathbf{v}_g$ when no collision is detected ($w\rightarrow 0$). As a safety mechanism if we detect $Z_\text{close}\le \tau_\text{safe}$, where $\tau_\text{safe}$ is a safe braking corridor, stop the quadrotor in its position to avoid collision and then move towards the free space. 

Finally, we adjust the baseline with respect to $Z_\text{close}$, i.e., $b = KZ_\text{close}$ (where $K>0$ is a tunable gain parameter). For implementation we use a low-pass filtered $Z_\text{close}$ to avoid matching artifacts.

The above method was tested both in simulated and real environments (Figs. \ref{fig:BlenderTopSceneForrest} and \ref{fig:Banner}{\color{red}a}). We can observe in Fig. \ref{fig:BlenderTopSceneForrest} that the quadrotor with large baseline crashes since the first tree is not visible and the quadrotor with smaller baseline can navigate to the goal but is 4$\times$ slower than the variable baseline quadrotor. 


\subsection{Flying through a static/dynamic unknown shaped gap}
\label{subsec:ExptsGapFlyt}
This scenario is inspired by the problem presented in GapFlyt\cite{GapFlyt}. A quadrotor is present in a scene equipped with a stereo camera setup whose baseline can be changed. The per-pixel depth `seen' by the camera can be modelled by a univariate bimodal Gaussian distribution. The contour $\mathcal{C}$ of the gap/opening (on the image plane) is defined as the pixels which have the maximum spatial depth disparity. Since, we have depth from disparity, this becomes a simple clustering problem. All the points which belong to the `far' cluster (average higher $Z$ value) are called background $\mathcal{B}$ and points in the `close' cluster are called foreground $\mathcal{F}$. Similar to \cite{GapFlyt}, we track both $\mathcal{F}$ and $\mathcal{B}$ separately (since its computationally cheaper and faster than running detection at every step) to infer $\mathcal{C}$. Contrary to \cite{GapFlyt}, when we lose track or for a dynamic gap, we re-detect the gap by clustering on the depth image. Our control policy is to align the center of the image with the safest point $\mathbf{x_s}$ defined as follows

\[
\mathbf{x}_s =
     \begin{cases}

     \mathbb{M}\left(\mathcal{F}\right) , &\quad \overline{\overline{\mathcal{F}}}\geq\overline{\overline{\mathcal{B}}} \\
       \mathbb{M}\left(\mathcal{B}\right), &\quad \text{otherwise}\\
     \end{cases}
\]
where $\mathbb{M}$ denotes the median operator.

Also, note that since we do not move the quadrotor to detect the gap, we can fly through dynamic gaps by invoking depth based gap detection (Fig. \ref{fig:Banner}{\color{red}b}). Again, we use a PID controller to align the center of the image with $\mathbf{x_s}$. 

Finally, we adjust the baseline with respect to our closeness to the gap, i.e., $b = K\mathbb{M}\left(Z_\mathcal{C}\right)$. Here, $Z_\mathcal{C}$ denotes the set of depth of all contour points and $K>0$ is a tunable gain parameter. Such adjustment of baseline makes depth estimation more accurate when the gap is near and also makes the quadrotor smaller making it easier to traverse the gap. 

We improve on speed of gap detection by $80\times$ (when using same computer for processing), speed of gap traversal by $20\%$ (3 ms$^{-1}$ over 2 ms$^{-1}$) and smallest gap clearance by $20\%$ (4 cm over 5 cm) when compared to \cite{GapFlyt}. We show our gap traversal sequences for both static and dynamic window in Figs. \ref{fig:GapFlytStatic} and \ref{fig:Banner}{\color{red}b} respectively. 

\subsection{Accurate IMO Detection}
\label{subsec:ExptsIMO}
Consider a scenario where we want to accurately detect the IMO in 3D either to dodge or pursue the target \cite{evdodgenet, van2018persistent}. First step is to detect the IMO then to track it. To make the problem simple, we assume that only one IMO is visible in the field of view. The IMO is first detected by clustering the tracklets' motion \cite{momswithevents} along with depth. Then the 3D trajectory of the IMO is tracked by computing depth from disparity. To improve accuracy of the estimated trajectory (especially in situations when the object is very close or very far from the camera) we adjust the baseline with respect to the median depth of the object, i.e.,  $b = K\mathbb{M}\left(Z_\text{IMO}\right)$. Here, $Z_\text{IMO}$ denotes the set of depth of all IMO points and $K>0$ is a tunable gain parameter. We show that using a variable baseline system we can track the object in a depth range larger than that just by using small/large baseline and with higher accuracy (Fig. \ref{fig:IMODiffBaselines}). Fig. \ref{fig:IMODiffBaselines} shows that the detection of the IMO fails in both small and large baseline cases when the object is very far or near to the camera but the variable baseline system can detect the object through the entire range.

\section{Conclusions}
\label{sec:Conc}
We applied the philosophy of morphable design to a stereo system to increase the depth sensing range. We discuss the hardware and software design of such a system for deployment on a mini-quadrotor. We present a simple calibration method and interpolate between the calibration values for the entire range using simple forward kinematics of the system. A through analysis of errors due to miscalibration is provided and we hope this will serve as a blueprint for researchers and practitioners alike. Finally, we demonstrate the utility of such a design philosophy in three different applications: (a) flying through a forest, (b) flying through a static/dynamic gap of unknown shape and location, and (c) accurate  3D pose detection of an independently moving object. We show that our variable baseline system is more accurate and robust in all the three scenarios when compared against a fixed baseline system. We feel that this would serve as a motivator for researchers and practitioners to incorporate sensors on controllable moving parts on robots to enable better perception capabilities.







\bibliographystyle{unsrt}
\bibliography{Ref}

\end{document}